# Unsupervised Multimodal Fusion of In-process Sensor Data for Advanced Manufacturing Process Monitoring

Matthew McKinney, Anthony Garland, Dale Cillessen, Jesse Adamczyk, Dan Bolintineanu, Michael Heiden, Elliott Fowler, Brad L. Boyce

Sandia National Laboratory

Highlights

- A novel unsupervised multimodal data fusion approach for manufacturing process monitoring is presented.
- Contrastive learning techniques correlate diverse sensor data without requiring labeled datasets.
- High-dimensional manufacturing data is compressed into low-dimensional representational spaces.
- This innovation facilitates data-driven decision-making for improved quality control and operational efficiency.

Abstract:

Effective monitoring of manufacturing processes is crucial for maintaining product quality and operational efficiency. Modern manufacturing environments often generate vast amounts of multimodal data, including visual imagery from various perspectives and resolutions, hyperspectral data, and machine health monitoring information such as actuator positions, accelerometer readings, and temperature measurements. However, interpreting this complex, high-dimensional data presents significant challenges, particularly when labeled datasets are unavailable or impractical to obtain. This paper presents a novel approach to multimodal sensor data fusion in manufacturing processes, inspired by the Contrastive Language-Image Pre-training (CLIP) model. We leverage contrastive learning techniques to correlate different data modalities without the need for labeled data, overcoming limitations of traditional supervised machine learning methods in manufacturing contexts. Our proposed method demonstrates the ability to handle and learn encoders for five distinct modalities: visual imagery, audio signals, laser position (x and y coordinates), and laser power measurements. By compressing these high-dimensional datasets into low-dimensional representational spaces, our approach facilitates downstream tasks such as process control, anomaly detection, and quality assurance. The unsupervised nature of our method makes it broadly applicable across various manufacturing domains, where large volumes of unlabeled sensor data are common. We evaluate the effectiveness of our approach through a series of experiments, demonstrating its potential to enhance process monitoring capabilities in advanced manufacturing systems. This research contributes to the field of smart manufacturing by providing a flexible, scalable framework for multimodal data fusion that can adapt to diverse manufacturing environments and sensor configurations. The proposed method paves the way for more robust, data-driven decision-making in complex manufacturing processes.

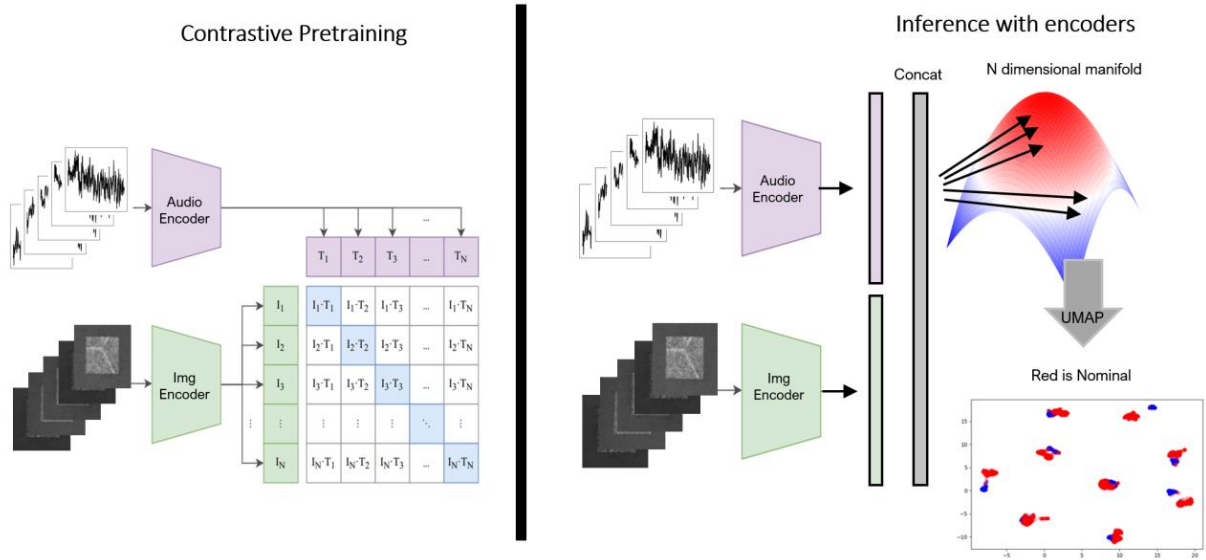

**Graphical Abstract** (left) We use contrastive loss to train encoders for each modality. Contrastive loss pushes corresponding vectors closer together in latent space (The blue diagonal shows $I_i T_j$, where $i = j$ ) while dissimilar vectors are pushed apart ($I_i T_j$ where $i \neq j$ ). (right) We use the encoders for inference over the data to identify clusters and anomalies. The red and blue dots on the 2D scatter plot are data tuples from a nominal print (in red) and a purposefully off-nominal print (blue). Each dot represents an individual part for a unique layer, and each group of red and blue circles represents a distinct part on the build plates. There were nine parts built. The red and blue dots are not directly on top of each other which shows we are able to discriminate between the nominal and off-nominal builds.

# 1 Background

## 1.1 Additive Manufacturing

Additive manufacturing (AM), particularly laser powder bed fusion (LPBF), has emerged as a transformative technology in advanced manufacturing. LPBF is a process where a high-power laser selectively melts and fuses metal powder particles layer by layer to create complex 3D objects (Kumar 2003). This technology offers unprecedented design freedom, material efficiency, rapid design-build-test cycles, and the ability to produce intricate geometries that are challenging or impossible with traditional manufacturing methods (Mohsen Seifi et al. 2016; White et al. 2021; Jared et al. 2017) . Figure 1 shows the internal view of the 3D Systems ProX DMP 200 LPBF printer used in this paper.

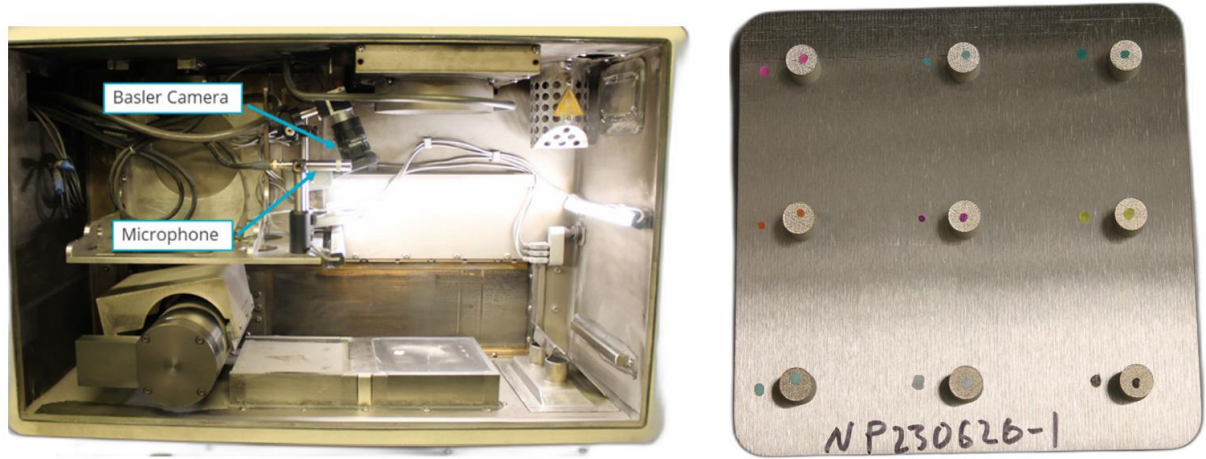

Figure 1 (left) ProX printer with camera and microphone. (right) Example build plate.

## 1.2   Machine Learning In Additive Manufacturing.

The adoption of machine learning (ML) techniques in AM has seen significant growth in recent years, driven by the need to address challenges in process optimization, quality control, and defect prediction (Zhang and Yan 2023; Song, Zhou, and Ahmed 2023). As an example, Gaikwad et al. (2020) demonstrated the use of ML for melt pool geometry prediction in LPBF, leveraging in-process thermal imaging data to improve process understanding (Gaikwad et al. 2020). Their work highlighted the potential of ML to capture complex process-structure-property relationships in AM. Defect detection and quality assurance remain critical challenges in AM. Westphal and Seitz (2022) proposed a machine learning approach for analyzing environmental sensor data to support quality assurance in fused deposition modeling, another AM technique (Westphal and Seitz 2022). Their work emphasized the importance of multimodal data integration for comprehensive process monitoring. In the context of LPBF, Zhang et al. developed a convolutional neural network (CNN) based method for detecting defects using melt pool, plume, and spatter information, demonstrating high accuracy in identifying build quality anomalies (Zhang et al. 2018). (Eschner et al. 2020) used audio signals and a neural network to predict density in parts thereby showing that ML can be effective for interpreting acoustic data in AM .Later, Zhiwei Zhang et al. used a super resolution algorithm and adapted YOLOv3 to detected lattice defects in LPBF thereby showing the benefit of chaining distinct ML algorithms together to achieve high quality outputs (Zhiwei Zhang et al. 2022).

The application of ML in AM extends beyond process monitoring to design optimization and parameter selection. Ko et al. (2021) introduced a framework combining machine learning and knowledge graphs for constructing design rules in AM (Ko et al. 2021). This approach aimed to capture and utilize the complex relationships between process parameters, part geometry, and final product quality, showcasing the potential of ML to enhance design for additive manufacturing (DfAM) practices. Garland et al. showed how to design pareto optimal lattice structures for AM using a machine learning algorithm to predict lattice performance and a genetic algorithm to iteratively design new structures (A. P. Garland et al. 2021). Brown et al. showed how to use reinforcement learning to tailor the force displacement response of lattice metamaterials to meet design objectives (N. K. Brown et al. 2023).

## 1.3 Current Methods In Multimodal Data Fusion Using Machine Learning

Despite these advancements, challenges remain in effectively utilizing the vast amounts of multimodal data generated during AM processing and throughout the full AM production lifecycle. While traditional statistical process control often employs data from individual, independent sources, the integration of diverse data sources, including visual imagery, audio signals, and machine parameters, offers the potential for a more comprehensive process understanding and better quality control (Grasso, Gallina, and Colosimo 2018; Zouhri et al. 2020; Jafari-Marandi et al. 2019; Gaikwad et al. 2020; Sarah K. Everton et al. 2016). However, effectively fusing and interpreting this high-dimensional, heterogeneous data presents significant challenges, particularly in the absence of large, labeled datasets (Zhiwei Zhang et al. 2022; Jia et al. 2024; Narayanan et al. 2020; Raihan et al. 2024).

Traditional approaches to sensor fusion in manufacturing have often relied on supervised machine learning techniques, requiring extensive labeled training data (Scime, Singh, and Paquit 2022). While effective in certain applications, these methods face limitations in scalability and adaptability to new manufacturing environments or process variations (Khanzadeh et al., 2018). Moreover, the time and cost associated with generating labeled datasets for each new application or process can be prohibitive in dynamic manufacturing settings (Narayanan et al. 2020; Ero, Taherkhani, and Toyserkani 2023).

One solution to generating labeled data is to use one modality of data to predict another using supervised machine learning approaches. Several studies have demonstrated the effectiveness of this approach. Khanzadeh et al. (2018) utilized thermal imaging data to predict porosity in direct laser deposition (Khanzadeh et al. 2018), while Snow et al. (2021) employed layerwise optical images to detect flaws identified by X-ray computed tomography in laser powder bed fusion (Snow et al. 2021). Hofmann et al. (2024) used infrared thermography data to predict porosity validated by X-ray micro-computed tomography in laser-based powder bed fusion of polyamide 12 (Hofmann et al. 2024). Similarly, (Pak et al. 2024) used sequences of thermal images to predict the location of pores using an adapted video vision transformer model. Estalaki et al. (2022) used in-process thermographic data to predict micropores in laser powder bed fusion materials (Estalaki et al. 2022). For the second modality, the ran simulations to predict the quality of the metal pool for every voxel in the part, and then used machine learning to predict the melt pool parameters from sensor data.

Further advancing the field, Grasso et al. (2018) explored data fusion methods for quality characterization in metal additive manufacturing, combining various sensor data used to control the printer to predict quality metrics (Grasso, Gallina, and Colosimo 2018; Mario Grasso et al. 2017). In the realm of ceramic additive manufacturing, Cheng et al. (2019) proposed a novel approach using photoacoustic signals to detect defects where the acoustic signal was matched to the laser position creating a photo-acoustic like image which was then used for defect detection. (Cheng, Lei, and Xiao 2019). These studies collectively demonstrate the potential of cross-modal machine learning techniques in enhancing process monitoring and quality assurance in additive manufacturing.

The integration of contrastive learning techniques, such as those inspired by CLIP (Alec Radford et al. 2021), into AM process monitoring represents a promising direction for addressing these challenges. By enabling the correlation of different data modalities without explicit labels, such approaches could potentially revolutionize defect detection, process optimization, and quality control in AM. However, the application of these advanced ML techniques to the specific challenges of AM, particularly in multimodal sensor fusion, remains an area ripe for further exploration and development.

## 1.4 Contrastive Learning and Multimodal fusion

Pretraining refers to the process of training a machine learning model on a large, general dataset to acquire a rich understanding of the underlying data structures and patterns (Chen et al. 2020). This initial training phase enables the model to develop a robust and generalizable representation of the data, which can then be leveraged as a foundation for various applications. In some cases, the pre-trained representations can be used directly for downstream tasks, without any modification or fine-tuning, as is the case with CLIP embeddings. This approach, known as "zero-shot" or "few-shot" learning, allows models to generalize to new tasks and domains with minimal additional training. By pretraining on a large dataset, models can develop a deep understanding of the data, which can be subsequently applied to a wide range of tasks and domains, thereby enhancing their performance and reducing the need for extensive task-specific training. The success of models like BERT (Devlin et al. 2019) in natural language processing demonstrated the effectiveness of pretraining on masked language modeling tasks. BERT like masks were latter extended to other modalities (Baevski et al. 2022), which suggest that similar techniques could extend to manufacturing data. In the visual domain, He et al. (2020) introduced MoCo (Momentum Contrast), which showed that contrastive learning could be used to pretrain visual representations that outperform supervised pretraining on various downstream tasks (Hu et al. 2021). The rise and expressiveness of modern large language models (GPT models) based on the self-supervised task of predicting the next token demonstrates the power of reframing traditional supervised learning paradigms as self-supervised next-token prediction tasks, enabling the development of highly expressive and generalizable models ("[PDF] Language Models Are Unsupervised Multitask Learners | Semantic Scholar," n.d.; T. B. Brown et al. 2020).

Contrastive loss is a type of method of pretraining that has proven effective for many problems. Notably, the Contrastive Language-Image Pre-training (CLIP) model developed by Radford et al. (Alec Radford et al. 2021) demonstrated the ability to learn powerful visual representations from natural language supervision without the need for traditional labeled datasets. CLIP used matching data pairs of images and captions. CLIP's simplicity and effectiveness for multimodal data fusion has resulted in the technique being widely used in the machine learning community - Radford et al's paper is among the most highly cited multimodal fusion papers to date. This approach has inspired new directions in multimodal learning across various domains (Junnan Li et al. 2023; 2022; Lin et al. 2024; Sahraoui et al. 2023).

Contrastive learning methods have gained particular attention due to their ability to learn discriminative features without explicit labels. The core idea behind contrastive learning is to train embeddings such that similar samples are pulled together in the embedding space while dissimilar samples are pushed apart. Chen et al. (2020) proposed SimCLR, a simple framework for contrastive learning of visual representations, which significantly improved the state-of-the-art in self-supervised and semi-supervised learning on ImageNet (Chen et al. 2020).

The integration of multiple modalities in contrastive learning frameworks has opened new avenues for representation learning. CLIP (Contrastive Language-Image Pretraining) by Radford et al. (2021) demonstrated the power of learning visual representations from natural language supervision, enabling zero-shot transfer to various vision tasks (Alec Radford et al. 2021). This approach has inspired numerous extensions and applications across different domains.

Recent advancements have further pushed the boundaries of contrastive and multimodal learning. Mustafa et al. introduced a model that combines contrastive learning with a mixture of experts architecture, enabling more efficient multimodal learning (Mustafa et al. 2022). The CoCa model extended the CLIP approach to enable both zero-shot image classification and open-ended image captioning, demonstrating the versatility of contrastive pretraining (Yu et al. 2022).

In the realm of self-supervised learning, Baevski et al. proposed a unified framework for learning representations across different modalities, showcasing the potential for more general-purpose multimodal models (Baevski et al. 2022). ImageBind further expanded on this idea, presenting a method for learning joint embeddings across six different modalities, including images, text, audio, depth, thermal, and IMU data (Girdhar et al. 2023).The success of these methods in learning powerful, transferable representations without extensive labeled data has significant implications for all domains of machine learning include generating meaningful representations from manufacturing data.

The Flamingo model demonstrated the potential of large-scale multimodal models to perform few-shot learning across a wide range of vision and language tasks, highlighting the adaptability of pretrained representations (Alayrac et al. 2022). This adaptability is particularly relevant for domains like manufacturing, where the ability to quickly adapt to new processes or products is crucial.

As these techniques continue to evolve, their potential applications in various fields, including manufacturing and process monitoring, are expanding. The ability to learn from unlabeled, multimodal data and adapt to new tasks with minimal fine-tuning aligns well with the needs of dynamic and complex manufacturing environments.

## 1.5 Contrastive Learning for Manufacturing

Our review so far of ML applications in Additive Manufacturing reveals both promising developments and significant challenges. First, ML clearly demonstrates the most promising potential to process in-situ monitoring data to help identify produce defects and process anomalies (Qin et al. 2022). However, current methodologies often struggle with effectively fusing multi-sensor data, particularly when dealing with high-dimensional inputs like audio, images, and thermal video. This challenge is evidenced by the scarcity of studies utilizing multi-modal data in AM literature. Despite the expectation that multimodal data would provide synergistic effects for better process monitoring and defect prediction, this is not clearly demonstrated in existing research. Furthermore, there's a significant gap in leveraging vast unlabeled data from manufacturing to deliver business or scientific value, a need that's particularly critical in modern manufacturing environments where enormous quantities of sensor data are generated daily.

Self-supervised machine learning in manufacturing remains limited, despite the widespread use of self-supervised approaches, such as contrastive learning, in other fields. Within AM, there's a pressing need for robust, scalable approaches to multimodal sensor fusion that can function effectively with limited or no labeled data. The manufacturing community often focuses on ML approaches that yield directly observable or actionable predictions, such as identifying specific defects. While this approach has its merits, it may inadvertently limit the range of ML techniques explored in the field. In contrast, the broader ML community is more open to methods that generate rich latent feature representations. These representations, while not immediately interpretable, often prove invaluable when applied creatively to various problems. This approach allows for more flexible and potentially powerful applications of ML, even when the initial utility of the learned representations isn't obvious. This difference in perspective highlights a potential area for growth in manufacturing ML applications. By embracing techniques that prioritize powerful representations over immediate interpretability, the manufacturing sector could unlock new insights and capabilities, potentially leading to breakthroughs in process understanding and optimization.

In this paper, we address these challenges by applying the CLIP (Contrastive Language-Image Pre-training) algorithm to Additive Manufacturing data. Our work demonstrates how this powerful self-supervised learning technique can be adapted to the manufacturing domain, potentially revolutionizing process monitoring, control capabilities, and defect prediction in AM. By leveraging CLIP's ability to create meaningful representations from diverse data modalities, we aim to unlock new insights and capabilities in the field of Additive Manufacturing.

## 2 Method

### 2.1 Dataset

We investigate two datasets: (1) an in-process dataset consisting of acoustic time-series recordings coupled with optical images, and (2) a post-process dataset consisting of several distinct imaging modalities of as-manufactured parts along with a measurement of product performance in the form of compressive force-displacement curves.

## 2.2 Additive manufacturing in process data

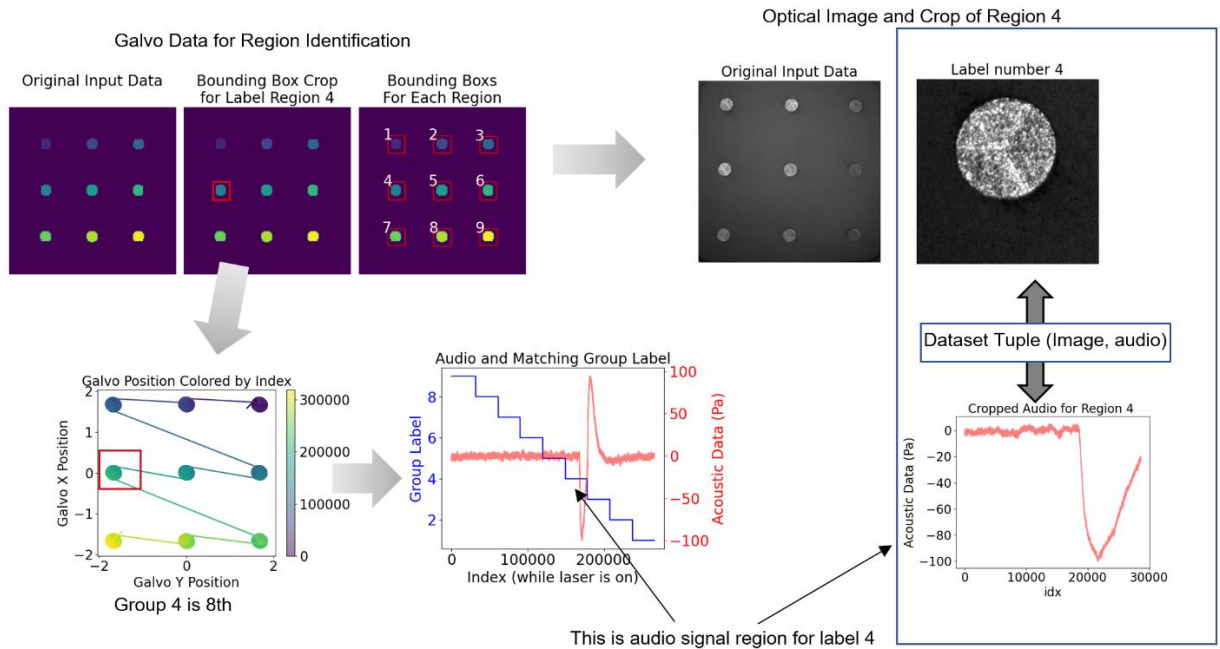

*Figure 2 Tuple generation from DAQ and Optical Data*

The first dataset (Figure 2) was developed at Sandia National Laboratories to explore anomaly detection in Laser Powder Bed Fusion (LPBF) processes. The dataset focuses on the manufacture of simple geometries: cubes and cylinders with dimensions of 10 x 10 x 10 mm, printed under a range of conditions encompassing both nominal and off-nominal configurations. Nominal prints were produced using 113-watt laser Power and 1400 mm/s laser speed on a laser (SPI Laser, JK300 FL) operating at a wavelength of 1064 nm, while off-nominal prints involved deviations from the nominal parameters.

The dataset comprises four distinct builds: nominal cubes, off-nominal cubes, nominal cylinders, and off-nominal cylinders. Each build consists of 9 printed items, with approximately 330 layers per build, resulting in about 11,000 correlated data tuples. The primary data modalities in this study are DAQ data and image data. A DAQ system collected audio, laser position, and laser power using a NI cDAQ-9178 with NI-9232 and NI-9223 data acquisition cards. Images were captured using a custom python-based process monitoring application that triggers a camera to capture an image after each layer is completed. The camera is inside the build chamber and is an optical camera (Basler ACA5472-17um) with a 100-mm working distance lens (Edmund Optics 16mm/F1.8 86-571, 43.2-degree field of view). The off-nominal cubes and off-nominal cylinder builds had 6 of the items printed with off-nominal print settings while 3 remained nominal (see Figure 3). These off-nominal printed items were printed to study how to best use in-process sensors to detected defects and process parameter drift.

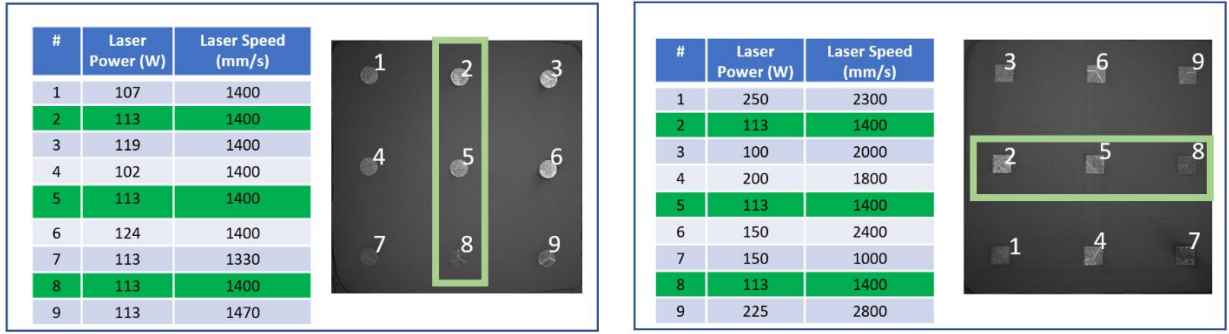

*Figure 3. Process parameters for the off-nominal cylinders (left) and off-nominal cubes (right) where the green rows and green box indicate the 3 items which are nominal.*

We primarily focused on audio data, as previous research indicates its ability to discriminate between different sounds and correlate them with specific defect types in LPBF processes (Kouprianoff et al. 2021; Eschner et al. 2020; Cheng, Lei, and Xiao 2019; Sun et al. 2024). The microphone (378A06 with Preamplifier 426E01 ICP 076539) is inside the build chamber. In later sections of this paper, we investigate using additional modalities, including the galvo mirror x and y positions (which direct the laser), laser on/off status, and the measured laser voltage.

The core challenge in processing this dataset lies in correlating the 1D audio signals with the 2D image data. To address this, we developed a workflow (illustrated in Figure 2) that involves identifying x, y positions where the laser is active, creating a 2D grid representation of laser activity, isolating individual regions using connected region analysis, labeling each connected region in the 2D space, identifying corresponding indices in the DAQ signal for each labeled region, cropping the audio signal at these indices, scaling the 2D grid to match image pixel coordinates, and finally cropping the layer-wise image accordingly.

This workflow enables us to isolate approximately 11,000 correlated data tuples, each consisting of a cropped image subregion and its corresponding audio signal (Figure 4). This multimodal approach, combining visual and audio data, along with additional process parameters, provides a comprehensive basis for anomaly detection in LPBF manufacturing.

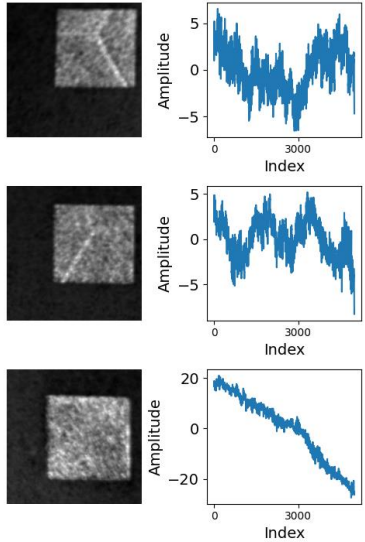

*Figure 4 Example of cropped image and cropped audio tuples.*

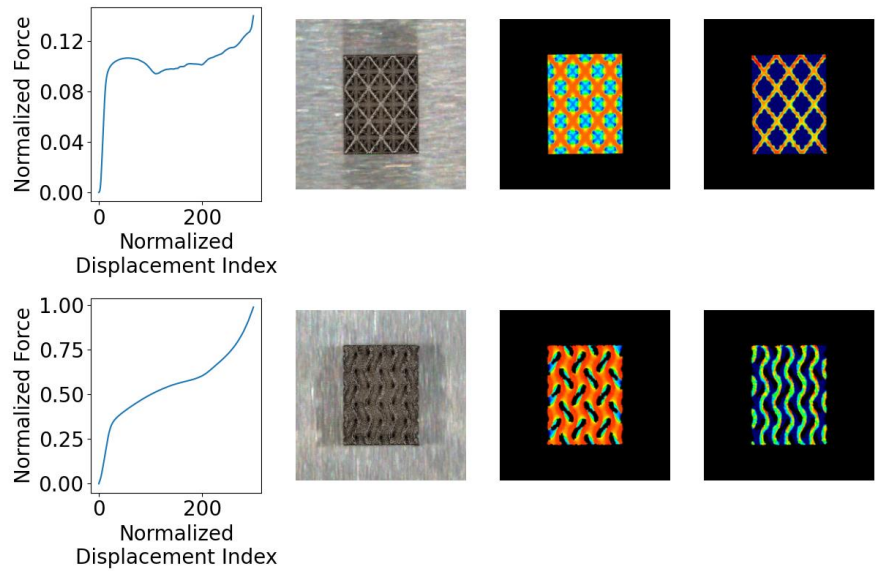

*Figure 5. Two examples of data tuples from the lattice dataset which includes normalized force-displacement curve, optical image, height map with large range, height map with range near the top surface.*

For our second dataset (Figure 5), we investigate learning on a much smaller dataset consisting of images and a performance response curve for compressive deformation. Our objective was to test the algorithm on a diverse set of modalities. The dataset is of additive manufactured lattices (octets and gyriods) which were imaged using three distinct imaging modalities and compressively deformed to measure their force-displacement response. The reader is referred to Garland et al. (2020) for further details on this dataset (Anthony Garland et al. 2020).

## 2.3 Machine Learning Architecture

Our model architecture is inspired by the Contrastive Language-Image Pre-training (CLIP) approach, adapted for the specific modalities present in our LPBF dataset. The core of our model consists of two primary encoders: an image encoder and an audio encoder, which are trained to project their respective inputs into a shared embedding space.

For the image encoder, we utilized a ResNet18 architecture (He et al. 2015). This choice was motivated by the ResNet architecture's proven effectiveness in various computer vision tasks and its ability to capture hierarchical visual features (Liu et al. 2022). However, the exact decision to use ResNet over a vision transformer is not important and both should work (Dosovitskiy et al. 2021). The image encoder model can be represented mathematically as a function

$$f_{img}: R^{H \times W \times C} \rightarrow R^D \qquad (1)$$

where H, W, and C are the height, width, and number of channels of the input image, respectively, and D is the dimension of the output embedding space.

Selecting an effective audio encoder was not obvious and we tested several different setups including MLPs and converting the audio to a spectrogram and then using a image encoder. In the end, we used the whisper model from OpenAI(Radford et al. 2022). The audio encoder leverages the tiny version of the Whisper model, which has demonstrated impressive performance in speech recognition tasks. The input to the whisper model is a log-mel-spectrogram with 80 channels. The model uses a positional embedding, convolutional layers, transformer layers, and Guassian error linear units (GELU) and is primarily a transformer mode. We modified the Whisper model by removing its classification head and adding two 1D convolutional layers to adapt the output to our desired embedding dimension. This encoder can be represented as a function

$$f_{audio}: R^T \rightarrow R^D, \qquad (2)$$

where T is the length of the input audio sequence.

Both encoders project their inputs into a shared 32-dimensional embedding space (D = 32). This relatively low-dimensional embedding space was chosen to encourage the model to learn compact, information-rich representations of the input data. Additionally, our model does not need to represent all natural images like the original CLIP model, but instead just in process data which is a significantly smaller. Similarly, the audio information has very strong prior of laser sintering metal powder and does not need to generalize to all audio inputs. This strong prior allows using a much smaller embedding space and relatively small encoder models to be used.

The model is trained using the CLIP contrastive loss function, which aims to maximize the cosine similarity between embeddings of matching image-audio pairs while minimizing the similarity between non-matching pairs. For a batch of N image-audio pairs, the loss function can be expressed as:

$$L = -\frac{1}{N}\sum_{i=1}^{N}\left[log\left(\frac{exp\left(\frac{sim(x_i,y_i)}{\tau}\right)}{\sum_{j=1}^{N}exp\left(\frac{sim(x_i,y_j)}{\tau}\right)}\right) + log\left(\frac{exp\left(\frac{sim(y_i,x_i)}{\tau}\right)}{\sum_{j=1}^{N}exp\left(\frac{sim(y_i,x_j)}{\tau}\right)}\right)\right] \quad (3)$$

Where $sim(u,v) = \frac{u^T v}{||u||||v||}$ is the cosine similarity between vectors u and v, τ is a temperature parameter that controls the softmax distribution, and $x_i$ and $y_i$ are the i-th image and audio inputs, respectively. For the lattice dataset where we have K encoders, we calculate the loss by computing the loss of each encoder against the other modalities (K - 1), and then summing these losses together.

After initial success, further discussed below, we added an additional encoder to our LPBF in-process setup to process the other DAQ signals. In addition to audio, the x and y position of the galvo mirror was collected and the voltage going to the laser was measured. Since this data is quite different than audio data a different approach was needed. We designed and implemented a MultiScaleLSTM as a novel neural network architecture designed to process time series data at multiple temporal scales simultaneously. It employs multiple LSTM heads, each operating on a different time scale of the input sequence (Hochreiter and Schmidhuber 1997). The model uses a power-of-two subsampling strategy for each head, effectively capturing both fine-grained and coarse-grained temporal patterns. Outputs from all heads are concatenated and passed through fully connected layers to produce the final output. This multi-scale approach allows the model to learn hierarchical temporal features, potentially improving its ability to capture both short-term and long-term dependencies in the data.

$$y = f(x) = W_2 \cdot ReLU(W_1 \cdot concat\left([h_i]_{\{i=0\}}^{\{N-1\}}\right) + b_1) + b_2 \quad (4)$$

where:

$$h_i = LSTM_i(x[::2^i])_T, i \in \{0, 1, \ldots, N-1\} \quad (5)$$

and x is the input sequence, N is the number of LSTM heads, $h_i$ is the final hidden state of the i-th LSTM head, $x[::2^i]$ denotes the input sequence subsampled by a factor of $2^i$, $LSTM_i(\cdot)_T$, represents the last output of the i-th LSTM, $W_1, W_2$ are weight matrices and $b_1, b_2$ are bias vectors of the fully connected layers, $concat[\cdot]$ represents the concatenation operation. The inspiration for the model was the multi-head attention in transformers networks (multiple LSTM paths) and the feature pyramid networks (the subsampling of the data) used in vision models which consider multiple resolutions of the image to identify objects at multiple scales. The pseudo code for the model is in the supplemental Figure 10.

## 2.4 Training

During training, we used a batch size of 8, which provided a good balance between computational efficiency and the number of contrastive pairs per batch. The relatively small batch size was chosen due to memory constraints and the high dimensionality of our input data, particularly the image modality.

The model was trained using the AdamW optimizer with a learning rate of 1e-4 and a weight decay of 1e-6 (Loshchilov and Hutter 2019). We employed the OneCycleLR learning rate schedule with a maximum learning rate of 3e-4 to gradually adjust the learning rate over the course of training, which helped in achieving better convergence (Smith and Topin 2018). We use gradient accumulation over multiple batches to help regularize the training.

To prevent overfitting and improve generalization, we applied various data augmentation techniques to both image and audio inputs. For image data, we utilized random cropping to a specified size, horizontal flipping, random rotation within a range of ±10 degrees, and color jittering to adjust the brightness, contrast, saturation, and hue. For audio data, we resampled the audio signal to a sample rate compatible with the Whisper model and cropped the audio to 30 seconds. If the signal was shorter than 30 seconds, we repeated it to fill the entire duration. These augmentations were crucial as they prevented the model from learning to identify data based on trivial cues such as the length of the audio signal, which could be associated with incorrect laser speed settings. We held out 5% of the data for validation to monitor the model's performance and generalization capabilities. The accuracy of the model was determined by taking the argmax of the product of the similarity scores and represents when the model can correctly predict which embedding vector goes with the matching vector from the other modality.

This multimodal contrastive learning approach enables our model to learn joint representations of the LPBF process that capture correlations between visual and auditory signals. These learned representations can then be used for downstream tasks such as anomaly detection, process monitoring, and quality control in additive manufacturing.

Figure 6 shows the accuracy, loss, and learning rate for the training of the CLIP model on the in-process dataset. We did not perform extensive hyper-parameter tunning since the combination of data, model, and training protocol were stable and showed good convergence.

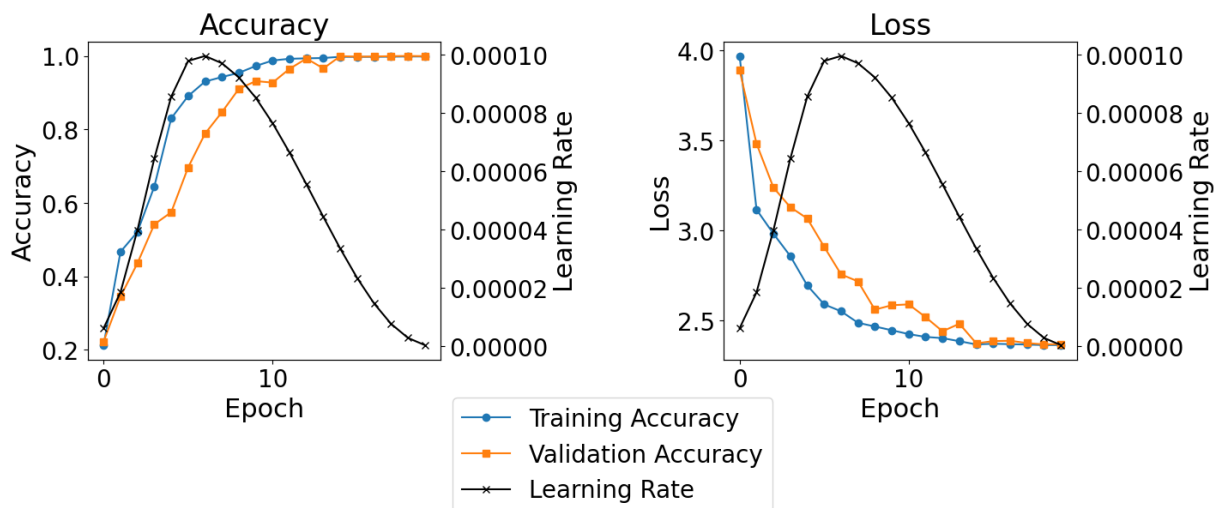

*Figure 6 Training CLIP. Optimization metrics for the in-process dataset.*

For the lattice dataset, we use three ResNet18 image encoders. For the force-displacement curve encoder we use a simple multi-layer perception model with skip. We use standard computer vision augmentation techniques on the optical images and height maps. For the force-displacement curve, we add noise to the data during training and random small shifts along the displacement dimension. These techniques help to regularize the training and prevent overfitting.

## 3 Results

We first evaluate the in-process dataset. After the encoders are trained, we pass all the data through each encoder, concatenate the vectors together, perform analysis to determine where the data tuples map to in the multi-dimensional space, and visualize the mapping as shown in the graphical abstract. We used Umap as our dimensionality reduction algorithm to visualize the latent (McInnes, Healy, and Melville 2020). The output embeddings are show in Figure 7.

Similarly, with the lattice builds, we passed all our data through the trained encoders. Using just the concatenated image vectors, we performed clustering of the latent space which is shown in the top right of Figure 8. Similarly, the bottom row of Figure 8, we passed all the data through the force-displacement encoder and then plotted the reduced latent space in the bottom right of Figure 8.

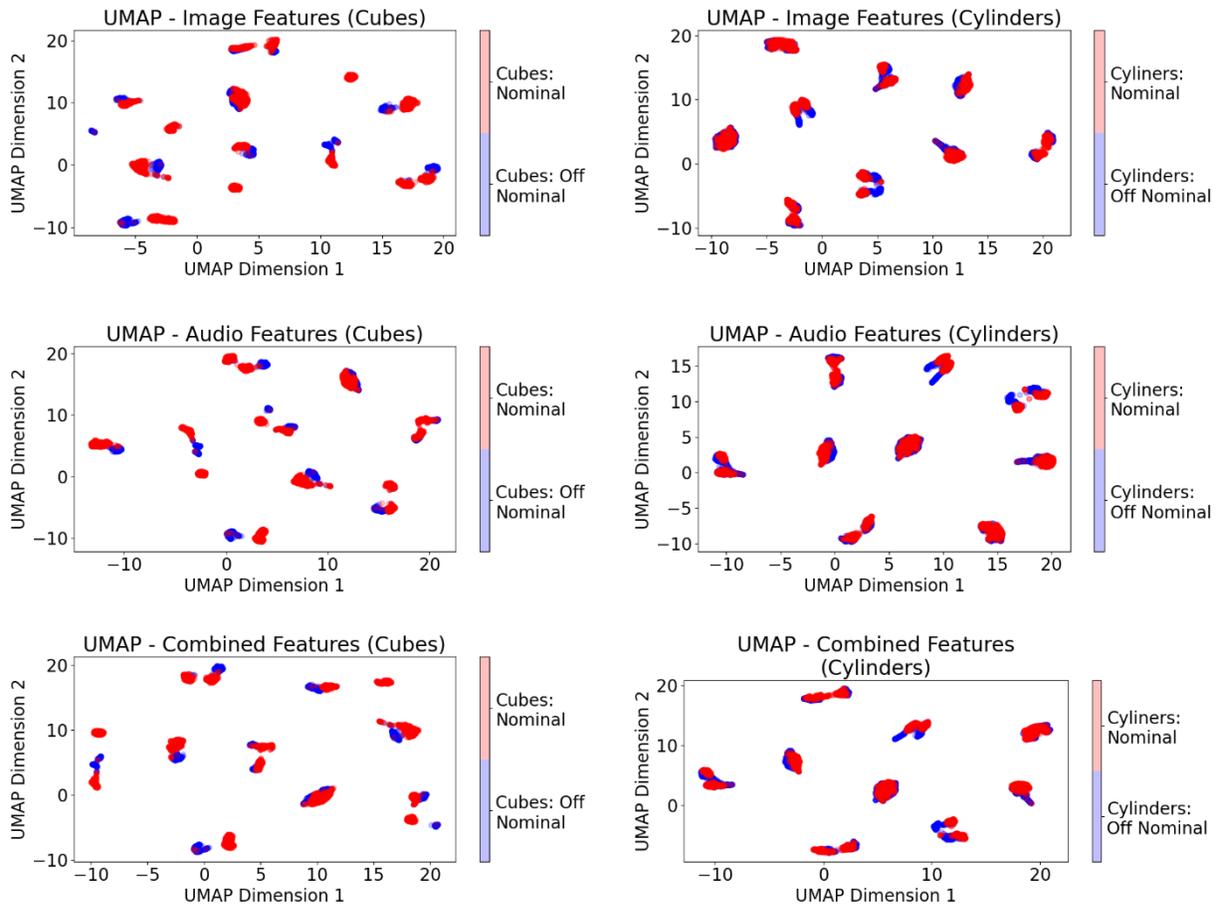

Figure 7 Umap dimensionality reduction of the embedding spaces for the image encoder, audio encoder, and the concatenated image and audio encoder. Left column shows results for the cube builds. Right column shows the result for the cylinder builds. Three of the items in the off-nominal build used nominal print settings.

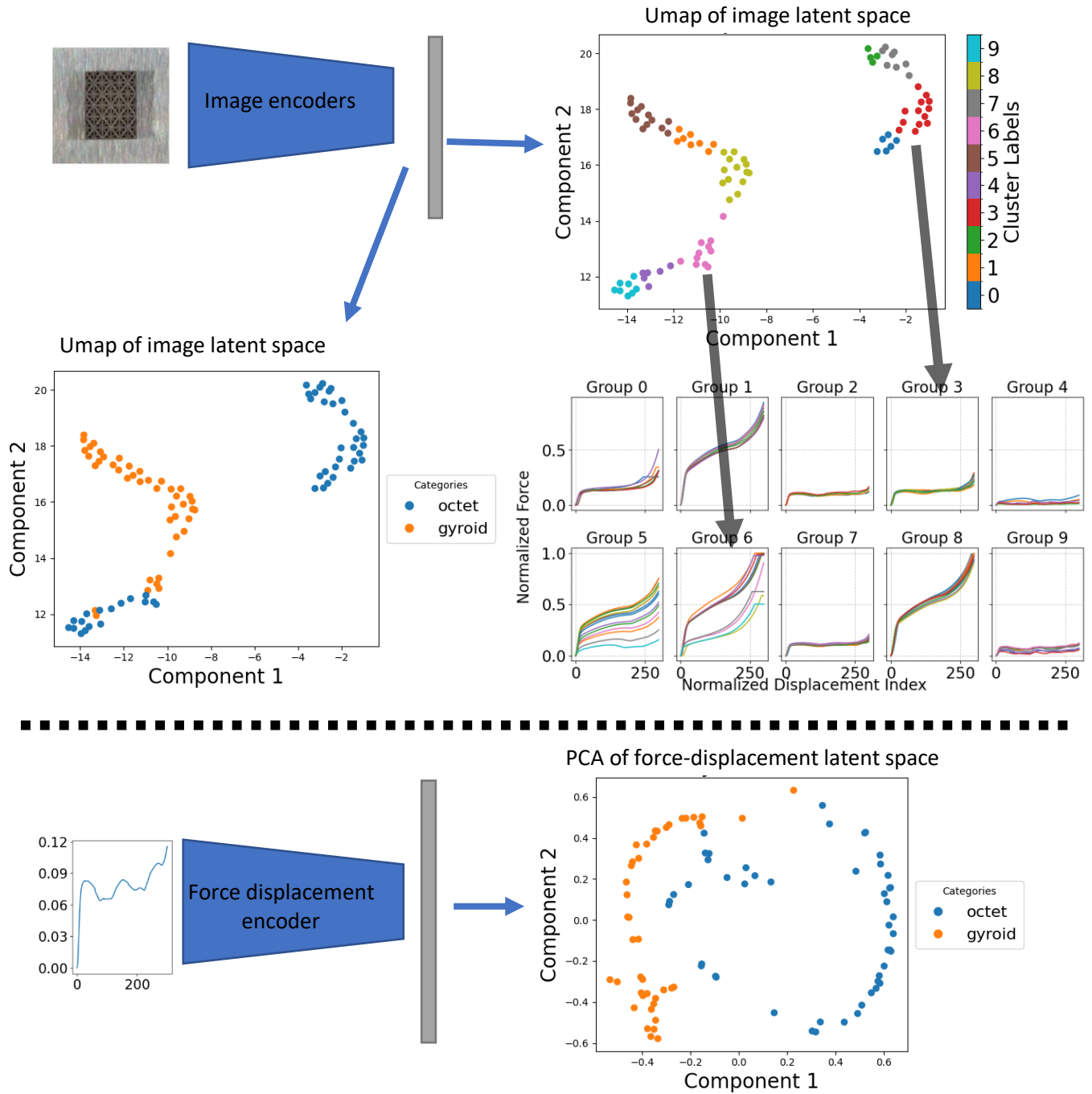

Figure 8. Latent space analysis of the lattice encoders. Top left indicates the images passing through the trained image encoders. The latent space is then clustered using kmeans and the clusters are show in the top right. The top right and middle left are the umap 2D representation of the latent space. The middle left shows that the latent space from the images colored by lattice type. The middle right shows the force displacement curves of each cluster where the clusters were generated from the image latent space. The bottom left indicates the force displacement curves being passed through the trained encoder. The bottom right shows the PCA reduction of the latent space colored by lattice type.

After seeing the good result from CLIP on the audio and vision modalities in the in-process dataset and seeing the good result form using multiple encoders from the lattices, we attempted to also encode the additional DAQ channel information. This information included the x and y position of the laser and the laser power output. We used the MultiScaleLSTM described earlier as the encoder. The training was similarly stable and reached 100% accuracy for the training and validation datasets. After training, we passed the data through the encoder to generate vectors and then used UMAP to convert the embeddings to 2D for plotting.

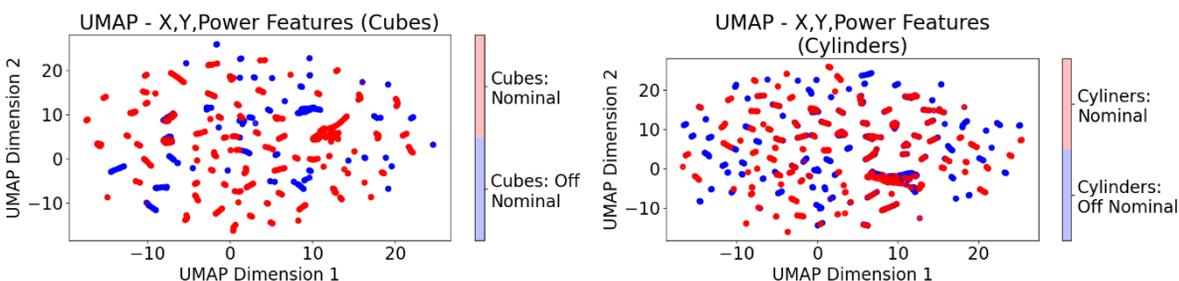

Figure 9. Embedding data from the x, y, power encoder for both the cubes (left) and cylinders (right).

## 4 Discussion

Our analysis of the cylinder and cube builds revealed intriguing insights into the effectiveness of our multimodal fusion approach. The emergence of nine distinct clusters in the latent space, which map to the 9 items printed, demonstrates the model's ability to discriminate the positional information on the build plate. These differences in position are not obvious (for example the left subfigure in Figure 4). This is particularly noteworthy given that we did not use pre-trained models, indicating that the encoders successfully learned meaningful representations from random initialization during the training process. The substantial variation observed in the latent space for cube builds is especially encouraging, as it suggests that significant changes in laser process parameters result in correspondingly large variations in the latent space. The variation in latent space can be used for anomaly detection using techniques like FADS (A. Garland, Potter, and Smith 2022) and PatchCore (Roth et al. 2022).

The success of our approach in encoding both image and audio data is significant. While using standard image encoders like ResNet18 is common practice, the effectiveness of our adapted Whisper model for processing additive manufacturing audio signals was not guaranteed. The model's ability to bridge the domain gap from human speech to manufacturing acoustics represents a notable achievement. Moreover, the encoders' capacity to learn meaningful representations despite the visual similarity of cubes and cylinders, and the potential for near-identical cropped images, underscores the robustness of our method.

However, our results also highlight areas for improvement, particularly in anomaly detection. The quality of input data, specifically from cameras and microphones, plays a crucial role in the model's performance. Our experiments revealed limitations in camera focus across the build plate, potentially hindering the detection of subtle textural differences resulting from variations in laser power or speed. This finding emphasizes the importance of using well-focused cameras covering the entire build plate in future work. Similarly, audio signal processing could benefit from higher-quality microphones and potentially multiple microphone setups, especially for multi-laser systems where triangulation of laser positions could provide additional valuable information.

While a subject matter expert might be able to look at the cropped images and identify changes from one image to the next, analyzing the raw audio signal is very difficult, as is can be seen in Figure 2 and Figure 4. Additionally, the large dip and spike in the audio signal in Figure 4 is the gas purge valve turning on and off. Interestingly, the audio-encoder has learned to ignore this signal.

The latent space analysis of the lattice encoders yielded intriguing results, demonstrating the power of our CLIP-inspired algorithm in generating meaningful representations. Clustering of the image encoder's latent space revealed distinct groups with characteristic force-displacement curves (Figure 8, middle right), indicating that the image data alone contains sufficient information to predict mechanical responses. This finding underscores the potential of our approach in bridging visual and mechanical properties of lattice structures. Conversely, the force-displacement encoder vectors exhibited a clear separation between octet and gyroid lattice data in the PCA-reduced space (Figure 8, bottom right). This emergent classification capability is particularly noteworthy, as it was not an explicit training objective. The mutual informativeness of these latent spaces—derived from image and force-displacement data, respectively—highlights the success of our training methodology in capturing and interrelating multiple modalities of lattice characteristics. These results suggest that our encoder framework has learned to extract and represent fundamental features that transcend the immediate training tasks, potentially offering new insights into the relationship between lattice geometry and mechanical behavior.

The encoded x, y, and power data shown in Figure 9 is more difficult to interpret. The encoder essentially learns how to take the laser tool-path and power for a part and map it to a reduced order space. The toolpath algorithm for two layers with exactly the same geometry is not the same since the computer aided manufacturing (CAM) software will rotate the raster direction of the laser path or perform a different raster strategy from one layer to the next to prevent anisotropic material properties. As a result, Figure 9 shows far more than 9 clusters. Each grouping represents a unique toolpath or laser power setting. A key and perhaps obvious result of Figure 9 is that the red and blue dots are not on top of each other, and therefore we can easily conclude that the builds were not the same and therefore we should not expect the same physical properties from the two builds. From a manufacturing process control perspective, this is an important result. However, we note that likely, you could process the data without using a deep neural network machine learning algorithm and come to a similar conclusion. More work is needed to understand how this additional encoder effects the training and quality of the image and audio encoders during the training protocol.

For all our training runs, we found the contrastive loss algorithm very stable and easy to train without extensive hyper parameter tuning. As a result, we did not perform extensive ablation studies on model architecture decisions or training protocols. Future works should investigate model and training protocol modifications.

# 5   Conclusion

Our unsupervised multimodal fusion approach for manufacturing process monitoring demonstrates significant promise in advancing the field of smart manufacturing. By successfully integrating diverse sensor data without the need for labeled datasets, we have developed a flexible and scalable framework applicable across various manufacturing domains. The method's ability to compress high-dimensional data into meaningful low-dimensional representations facilitates improved process monitoring, anomaly detection, and quality control.

We show the ability to take a relatively high dimension and high cost measurement (images) and correlated it a low dimension and low cost measurement (audio). This correlation of high cost and low-cost measurements by using machine learning has broad applications for engineering and scientific data analysis where it might be impractical to use a high cost instrument in some scenarios, but a machine learning correlated measurement (as demonstrated in this paper) from a cheaper system could be used in its place.

The physical interpretability of our results, particularly in the analysis of lattice structures, opens new avenues for discovering hidden insights in manufacturing processes. This capability could lead to optimized designs, improved process parameters, and a deeper understanding of the relationships between manufacturing inputs and material properties.

While we did not investigate model and training protocol modifications, a future work could investigate what is the minimum viable encoder for each modality. Smaller models could perform inference in real time and potentially enable real-time process control. We note that our setup for with ResNet18 and whisper 'tiny' is already quite fast for post layer analysis. Using an Nvidia, A6000 GPU on a single workstation computer we achieved inference speed of greater than 90 data tuples pers second. While real time control has had success in other metal additive manufacturing processes (like Wire Arc Additive Manufacturing), in LPBF the laser is moving at a much faster speed and would require machine learning algorithms to perform inference in the kilo-hertz range (Mattera, Nele, and Paolella 2024). More work is needed to make the encoder models highly efficient and run extremely fast on embedded systems to enable real time control.

While our study also revealed areas for improvement, such as the need for higher-quality sensor data and more focused imaging systems, these findings provide clear directions for future research. Addressing these limitations could further enhance the accuracy and reliability of our approach. Additionally, our cubes and cylinders are quite simplistic and extending the work in this paper to actual production parts that have complex geometries is needed.  Other modalities should also be investigated in the future, such as photodiode, inferred cameras, and CT voxel information. This could lead to mapping the learned representations to specific types of defects.

In conclusion, this work represents a significant step forward in leveraging machine learning for advanced manufacturing process monitoring. By enabling data-driven decision-making and providing a framework for interpreting complex, multimodal data, our method has the potential to drive innovations in quality control, process optimization, and material design in additive manufacturing and beyond.

# 6   Acknowledgement

Sandia National Laboratories is a multi-mission laboratory managed and operated by National Technology & Engineering Solutions of Sandia, LLC (NTESS), a wholly owned subsidiary of Honeywell International Inc., for the U.S. Department of Energy's National Nuclear Security Administration (DOE/NNSA) under contract DE-NA0003525. This written work is authored by an employee of NTESS.

The employee, not NTESS, owns the right, title and interest in and to the written work and is responsible for its contents. Any subjective views or opinions that might be expressed in the written work do not necessarily represent the views of the U.S. Government. The publisher acknowledges that the U.S. Government retains a non-exclusive, paid-up, irrevocable, world-wide license to publish or reproduce the published form of this written work or allow others to do so, for U.S. Government purposes. The DOE will provide public access to results of federally sponsored research in accordance with the DOE Public Access Plan.

# 7 References

Alayrac, Jean-Baptiste, Jeff Donahue, Pauline Luc, Antoine Miech, Iain Barr, Yana Hasson, Karel Lenc, et al. 2022. "Flamingo: A Visual Language Model for Few-Shot Learning." arXiv. https://doi.org/10.48550/arXiv.2204.14198.

Alec Radford, Jong Wook Kim, Chris Hallacy, A. Ramesh, Gabriel Goh, Sandhini Agarwal, Girish Sastry, et al. 2021. "Learning Transferable Visual Models From Natural Language Supervision." *International Conference on Machine Learning*.

Anthony Garland, Anthony Garland, Benjamin White, Benjamin C. White, Bradley Howell Jared, Bradley H. Jared, Michael Heiden, et al. 2020. "Deep Convolutional Neural Networks as a Rapid Screening Tool for Complex Additively Manufactured Structures." *Additive Manufacturing* 35 (October):101217. https://doi.org/10.1016/j.addma.2020.101217.

Baevski, Alexei, Wei-Ning Hsu, Qiantong Xu, Arun Babu, Jiatao Gu, and Michael Auli. 2022. "Data2vec: A General Framework for Self-Supervised Learning in Speech, Vision and Language." arXiv. https://doi.org/10.48550/arXiv.2202.03555.

Brown, Nathan K., Anthony P. Garland, Georges M. Fadel, and Gang Li. 2023. "Deep Reinforcement Learning for the Rapid On-Demand Design of Mechanical Metamaterials with Targeted Nonlinear Deformation Responses." *Engineering Applications of Artificial Intelligence* 126 (November):106998. https://doi.org/10.1016/j.engappai.2023.106998.

Brown, Tom B., Benjamin Mann, Nick Ryder, Melanie Subbiah, Jared Kaplan, Prafulla Dhariwal, Arvind Neelakantan, et al. 2020. "Language Models Are Few-Shot Learners." arXiv. https://doi.org/10.48550/arXiv.2005.14165.

Chen, Ting, Simon Kornblith, Mohammad Norouzi, and Geoffrey Hinton. 2020. "A Simple Framework for Contrastive Learning of Visual Representations." arXiv. https://doi.org/10.48550/arXiv.2002.05709.

Cheng, Baokai, Jincheng Lei, and Hai Xiao. 2019. "A Photoacoustic Imaging Method for In-Situ Monitoring of Laser Assisted Ceramic Additive Manufacturing." *Optics & Laser Technology* 115 (July):459–64. https://doi.org/10.1016/j.optlastec.2019.02.055.

Devlin, Jacob, Ming-Wei Chang, Kenton Lee, and Kristina Toutanova. 2019. "BERT: Pre-Training of Deep Bidirectional Transformers for Language Understanding." arXiv. https://doi.org/10.48550/arXiv.1810.04805.

Dosovitskiy, Alexey, Lucas Beyer, Alexander Kolesnikov, Dirk Weissenborn, Xiaohua Zhai, Thomas Unterthiner, Mostafa Dehghani, et al. 2021. "An Image Is Worth 16x16 Words: Transformers for Image Recognition at Scale." arXiv. https://doi.org/10.48550/arXiv.2010.11929.

Ero, Osazee, Katayoon Taherkhani, and Ehsan Toyserkani. 2023. "Optical Tomography and Machine Learning for In-Situ Defects Detection in Laser Powder Bed Fusion: A Self-Organizing Map and U-Net Based Approach." *Additive Manufacturing* 78 (September):103894. https://doi.org/10.1016/j.addma.2023.103894.

Eschner, N., L. Weiser, B. Häfner, and G. Lanza. 2020. "Classification of Specimen Density in Laser Powder Bed Fusion (L-PBF) Using in-Process Structure-Borne Acoustic Process Emissions." *Additive Manufacturing* 34 (August):101324. https://doi.org/10.1016/j.addma.2020.101324.

Estalaki, Sina Malakpour, Cody S. Lough, Robert G. Landers, Edward C. Kinzel, and Tengfei Luo. 2022. "Predicting Defects in Laser Powder Bed Fusion Using *in-Situ* Thermal Imaging Data and Machine Learning." *Additive Manufacturing* 58 (October):103008. https://doi.org/10.1016/j.addma.2022.103008.

Gaikwad, Aniruddha, Brian Giera, Gabriel M. Guss, Jean-Baptiste Forien, Manyalibo J. Matthews, and Prahalada Rao. 2020. "Heterogeneous Sensing and Scientific Machine Learning for Quality Assurance in Laser Powder Bed Fusion – A Single-Track Study." *Additive Manufacturing* 36 (December):101659. https://doi.org/10.1016/j.addma.2020.101659.

Garland, Anthony P., Benjamin C. White, Scott C. Jensen, and Brad L. Boyce. 2021. "Pragmatic Generative Optimization of Novel Structural Lattice Metamaterials with Machine Learning." *Materials & Design* 203 (May):109632. https://doi.org/10.1016/j.matdes.2021.109632.

Garland, Anthony, Kevin Potter, and Matt Smith. 2022. "Feature Anomaly Detection System (FADS) for Intelligent Manufacturing." arXiv. https://doi.org/10.48550/arXiv.2204.10318.

Girdhar, Rohit, Alaaeldin El-Nouby, Zhuang Liu, Mannat Singh, Kalyan Vasudev Alwala, Armand Joulin, and Ishan Misra. 2023. "ImageBind: One Embedding Space To Bind Them All." arXiv. https://doi.org/10.48550/arXiv.2305.05665.

Grasso, Marco, Francesco Gallina, and Bianca Maria Colosimo. 2018. "Data Fusion Methods for Statistical Process Monitoring and Quality Characterization in Metal Additive Manufacturing." *Procedia CIRP*, The 15th CIRP Conference on Computer Aided Tolerancing, CIRP CAT 2018, 11-13 June 2018, Milan, Italy, 75 (January):103–7. https://doi.org/10.1016/j.procir.2018.04.045.

He, Kaiming, Xiangyu Zhang, Shaoqing Ren, and Jian Sun. 2015. "Deep Residual Learning for Image Recognition." arXiv. https://doi.org/10.48550/arXiv.1512.03385.

Hochreiter, Sepp, and Jürgen Schmidhuber. 1997. "Long Short-Term Memory." *Neural Computation* 9 (8): 1735–80. https://doi.org/10.1162/neco.1997.9.8.1735.

Hofmann, Joseph, Ziqi Li, Kirsten Taphorn, Julia Herzen, and Katrin Wudy. 2024. "Porosity Prediction in Laser-Based Powder Bed Fusion of Polyamide 12 Using Infrared Thermography and Machine Learning." *Additive Manufacturing* 85 (April):104176. https://doi.org/10.1016/j.addma.2024.104176.

Hu, Peng, Xi Peng, Hongyuan Zhu, Liangli Zhen, and Jie Lin. 2021. "Learning Cross-Modal Retrieval with Noisy Labels." In *2021 IEEE/CVF Conference on Computer Vision and Pattern Recognition (CVPR)*, 5399–5409. Nashville, TN, USA: IEEE. https://doi.org/10.1109/CVPR46437.2021.00536.

Jafari-Marandi, Ruholla, Mojtaba Khanzadeh, Wenmeng Tian, Brian Smith, and Linkan Bian. 2019. "From *in-Situ* Monitoring toward High-Throughput Process Control: Cost-Driven Decision-Making Framework for Laser-Based Additive Manufacturing." *Journal of Manufacturing Systems* 51 (April):29–41. https://doi.org/10.1016/j.jmsy.2019.02.005.

Jared, Bradley H., Miguel A. Aguilo, Lauren L. Beghini, Brad L. Boyce, Brett W. Clark, Adam Cook, Bryan J. Kaehr, and Joshua Robbins. 2017. "Additive Manufacturing: Toward Holistic Design." *Scripta Materialia* 135 (July):141–47. https://doi.org/10.1016/j.scriptamat.2017.02.029.

Jia, Xinjian, Shan Li, Tongcai Wang, Bingshan Liu, Congcong Cui, Wei Li, and Gong Wang. 2024. "High-Performance Defect Detection Methods for Real-Time Monitoring of Ceramic Additive Manufacturing Process Based on Small-Scale Datasets." *Processes* 12 (4): 633. https://doi.org/10.3390/pr12040633.

Junnan Li, Dongxu Li, Caiming Xiong, and S. Hoi. 2022. "BLIP: Bootstrapping Language-Image Pre-Training for Unified Vision-Language Understanding and Generation." *International Conference on Machine Learning*.

Junnan Li, Dongxu Li, Silvio Savarese, and Steven C. H. Hoi. 2023. "BLIP-2: Bootstrapping Language-Image Pre-Training with Frozen Image Encoders and Large Language Models." *International Conference on Machine Learning*, January. https://doi.org/10.48550/arxiv.2301.12597.

Khanzadeh, Mojtaba, Sudipta Chowdhury, Mohammad Marufuzzaman, Mark A. Tschopp, and Linkan Bian. 2018. "Porosity Prediction: Supervised-Learning of Thermal History for Direct Laser Deposition." *Journal of Manufacturing Systems* 47 (April):69–82. https://doi.org/10.1016/j.jmsy.2018.04.001.

Ko, Hyunwoong, Paul Witherell, Yan Lu, Samyeon Kim, and David W. Rosen. 2021. "Machine Learning and Knowledge Graph Based Design Rule Construction for Additive Manufacturing." *Additive Manufacturing* 37 (January):101620. https://doi.org/10.1016/j.addma.2020.101620.

Kouprianoff, D., I. Yadroitsava, A. du Plessis, N. Luwes, and I. Yadroitsev. 2021. "Monitoring of Laser Powder Bed Fusion by Acoustic Emission: Investigation of Single Tracks and Layers." *Frontiers in Mechanical Engineering* 7 (June). https://doi.org/10.3389/fmech.2021.678076.

Kumar, Sanjay. 2003. "Selective Laser Sintering: A Qualitative and Objective Approach." *JOM* 55 (10): 43–47. https://doi.org/10.1007/s11837-003-0175-y.

Lin, Shuyang, Tong Jia, Hao Wang, Bowen Ma, Mingyuan Li, and Dongyue Chen. 2024. "Open-Vocabulary X-Ray Prohibited Item Detection via Fine-Tuning CLIP." arXiv. https://doi.org/10.48550/arXiv.2406.10961.

Liu, Zhuang, Hanzi Mao, Chao-Yuan Wu, Christoph Feichtenhofer, Trevor Darrell, and Saining Xie. 2022. "A ConvNet for the 2020s." arXiv. https://doi.org/10.48550/arXiv.2201.03545.

Loshchilov, Ilya, and Frank Hutter. 2019. "Decoupled Weight Decay Regularization." arXiv. https://doi.org/10.48550/arXiv.1711.05101.

Mario Grasso, Marco Grasso, Vittorio Laguzza, Vittorio Laguzza, Quirico Semeraro, Quirico Semeraro, Bianca Maria Colosimo, and Bianca Maria Colosimo. 2017. "In-Process Monitoring of Selective Laser Melting: Spatial Detection of Defects via Image Data Analysis." *Journal of Manufacturing Science and Engineering-Transactions of The Asme* 139 (5): 051001. https://doi.org/10.1115/1.4034715.

Mattera, Giulio, Luigi Nele, and Davide Paolella. 2024. "Monitoring and Control the Wire Arc Additive Manufacturing Process Using Artificial Intelligence Techniques: A Review." *Journal of Intelligent Manufacturing* 35 (2): 467–97. https://doi.org/10.1007/s10845-023-02085-5.

McInnes, Leland, John Healy, and James Melville. 2020. "UMAP: Uniform Manifold Approximation and Projection for Dimension Reduction." arXiv. https://doi.org/10.48550/arXiv.1802.03426.

Mohsen Seifi, Mohsen Seifi, Ayman A. Salem, Ayman A. Salem, Jack Beuth, Jack Beuth, Ola L. A. Harrysson, Ola L. A. Harrysson, John J. Lewandowski, and John J. Lewandowski. 2016. "Overview of Materials Qualification Needs for Metal Additive Manufacturing." *JOM* 68 (3): 747–64. https://doi.org/10.1007/s11837-015-1810-0.

Mustafa, Basil, Carlos Riquelme, Joan Puigcerver, Rodolphe Jenatton, and Neil Houlsby. 2022. "Multimodal Contrastive Learning with LIMoE: The Language-Image Mixture of Experts." arXiv. https://doi.org/10.48550/arXiv.2206.02770.

Narayanan, Barath Narayanan, Kelly Beigh, Solomon Duning, and Dathan Erdahl. 2020. "Material Identification and Segmentation Using Deep Learning for Laser Powder Bed Fusion." In *Applications of Machine Learning 2020*, edited by Michael E. Zelinski, Tarek M. Taha, Jonathan Howe, Abdul A. Awwal, and Khan M. Iftekharuddin, 25. Online Only, United States: SPIE. https://doi.org/10.1117/12.2567007.

Pak, Peter Myung-Won, Francis Ogoke, Andrew Polonsky, Anthony Garland, Dan S. Bolintineanu, Dan R. Moser, Michael J. Heiden, and Amir Barati Farimani. 2024. "ThermoPore: Predicting Part Porosity Based on Thermal Images Using Deep Learning." arXiv. https://doi.org/10.48550/arXiv.2404.16882.

"[PDF] Language Models Are Unsupervised Multitask Learners | Semantic Scholar." n.d. Accessed August 9, 2024. https://www.semanticscholar.org/paper/Language-Models-are-Unsupervised-Multitask-Learners-Radford-Wu/9405cc0d6169988371b2755e573cc28650d14dfe.

Qin, Jian, Fu Hu, Ying Liu, Paul Witherell, Charlie C. L. Wang, David W. Rosen, Timothy W. Simpson, Yan Lu, and Qian Tang. 2022. "Research and Application of Machine Learning for Additive Manufacturing." *Additive Manufacturing* 52 (April):102691. https://doi.org/10.1016/j.addma.2022.102691.

Radford, Alec, Jong Wook Kim, Tao Xu, Greg Brockman, Christine McLeavey, and Ilya Sutskever. 2022. "Robust Speech Recognition via Large-Scale Weak Supervision." arXiv. https://doi.org/10.48550/arXiv.2212.04356.

Raihan, Ahmed Shoyeb, Hamed Khosravi, Tanveer Hossain Bhuiyan, and Imtiaz Ahmed. 2024. "An Augmented Surprise-Guided Sequential Learning Framework for Predicting the Melt Pool Geometry." *Journal of Manufacturing Systems* 75 (August):56–77. https://doi.org/10.1016/j.jmsy.2024.05.023.

Roth, Karsten, Latha Pemula, Joaquin Zepeda, Bernhard Schölkopf, Thomas Brox, and Peter Gehler. 2022. "Towards Total Recall in Industrial Anomaly Detection." arXiv. https://doi.org/10.48550/arXiv.2106.08265.

Sahraoui, Maya, Youcef Sklab, Marc Pignal, Régine Vignes Lebbe, and Vincent Guigue. 2023. "Leveraging Multimodality for Biodiversity Data: Exploring Joint Representations of Species Descriptions and Specimen Images Using CLIP." *Biodiversity Information Science and Standards* 7 (September):e112666. https://doi.org/10.3897/biss.7.112666.

Sarah K. Everton, Sarah K. Everton, M. Hirsch, Matthias Hirsch, Petros Stravroulakis, Petros Stravroulakis, Richard Leach, Richard Leach, Adam T. Clare, and Adam T. Clare. 2016. "Review of In-Situ Process Monitoring and in-Situ Metrology for Metal Additive Manufacturing." *Materials & Design* 95 (April):431–45. https://doi.org/10.1016/j.matdes.2016.01.099.

Scime, Luke, Alka Singh, and Vincent Paquit. 2022. "A Scalable Digital Platform for the Use of Digital Twins in Additive Manufacturing." *Manufacturing Letters* 31 (January):28–32. https://doi.org/10.1016/j.mfglet.2021.05.007.

Smith, Leslie N., and Nicholay Topin. 2018. "Super-Convergence: Very Fast Training of Neural Networks Using Large Learning Rates." arXiv. https://doi.org/10.48550/arXiv.1708.07120.

Snow, Zackary, Brett Diehl, Edward W. Reutzel, and Abdalla Nassar. 2021. "Toward In-Situ Flaw Detection in Laser Powder Bed Fusion Additive Manufacturing through Layerwise Imagery and Machine Learning." *Journal of Manufacturing Systems* 59 (April):12–26. https://doi.org/10.1016/j.jmsy.2021.01.008.

Song, Binyang, Rui Zhou, and Faez Ahmed. 2023. "Multi-Modal Machine Learning in Engineering Design: A Review and Future Directions." *Journal of Computing and Information Science in Engineering* 24 (010801). https://doi.org/10.1115/1.4063954.

Sun, Yuchen, Sanam Gorgannejad, Aiden Martin, Jenny Nicolino, Maria Strantza, Jean-Baptiste Forien, Vivek Thampy, et al. 2024. "Direct Mechanistic Connection between Acoustic Signals and Melt Pool Morphology during Laser Powder Bed Fusion." *Applied Physics Letters* 125 (3): 034102. https://doi.org/10.1063/5.0205663.

Westphal, Erik, and Hermann Seitz. 2022. "Machine Learning for the Intelligent Analysis of 3D Printing Conditions Using Environmental Sensor Data to Support Quality Assurance." *Additive Manufacturing* 50 (February):102535. https://doi.org/10.1016/j.addma.2021.102535.

White, Benjamin C., Anthony Garland, Ryan Alberdi, and Brad L. Boyce. 2021. "Interpenetrating Lattices with Enhanced Mechanical Functionality." *Additive Manufacturing* 38 (February):101741. https://doi.org/10.1016/j.addma.2020.101741.